\documentclass[runningheads]{llncs}

\usepackage[T1]{fontenc}
\usepackage{graphicx,verbatim}
\usepackage{epsfig}
\usepackage{amsmath}
\usepackage{mathtools}
\usepackage{amsfonts}       % blackboard math symbols

\usepackage[]{hyperref}

\usepackage[nameinlink,capitalize]{cleveref}
\usepackage{booktabs}       % professional-quality tables
\usepackage{subfigure}
\usepackage{algorithm}
\usepackage{algpseudocode}
\usepackage{soul}
\usepackage{ragged2e}
\usepackage[usenames,dvipsnames]{xcolor}
\usepackage{microtype}

\usepackage{multirow}
\usepackage{colortbl}
\PassOptionsToPackage{table}{xcolor}
\usepackage{siunitx}
\usepackage{caption}

\definecolor{softgreen}{RGB}{220, 240, 220}
\definecolor{softred}{RGB}{250, 210, 210}

\newcommand{\githubrepo}{\url{https://github.com/ChristianGappGit/SSL_Pretraining}}

\usepackage{color}

\begin{document}
\title{Enabling Vision and Cross-Modal Learning for Multimodal Stroke Recurrence Prediction: An Interpretable Two-Step Framework}
\titlerunning{An Interpretable Two-Step Framework for Stroke Recurrence Prediction}
%\begin{comment}  %% Removed for anonymized MICCAI submission
\author{Christian Gapp\textsuperscript{\textbf{*}}\inst{1}\orcidID{0000-0002-4520-298X} 
	\and
	Elias Tappeiner\inst{1}\orcidID{0000-0003-1034-8361}
	\and
	Martin Welk\inst{1}\orcidID{0000-0002-6268-7050} 
	\and 
	Karl Fritscher\inst{2}\orcidID{0000-0003-2593-6203} 
	\and
	Stephanie Mangesius\inst{3,5}\orcidID{0000-0001-5790-2724}
	\and
	Constantin Eisenschink\inst{3,5}\orcidID{0009-0001-4845-8315}
	\and
	Philipp Deisl\inst{2,3,5}\orcidID{0009-0000-2228-0398}
	\and
	Michael Knoflach\inst{2,4}\orcidID{0000-0001-5576-6562}
	\and
	Astrid E. Grams \inst{3,5}\orcidID{0000-0003-2304-3946}
	\and
	Elke R. Gizewski\inst{3,5}\orcidID{0000-0001-6859-8377} 
	\and
	Rainer Schubert\inst{1}\orcidID{0000-0002-8026-7500} 
}
\authorrunning{C. Gapp et al.} % abbreviated author list (for running head)
\institute{Institute of Biomedical Image Analysis\\
	UMIT TIROL -- Private University for Health Sciences and Health Technology,	Eduard-Walln\"ofer-Zentrum 1, 6060 Hall in Tirol, Austria \and 
	VASCage -- Centre on Clinical Stroke Research, 6020 Innsbruck, Austria \and
	Department of Radiology, \and Department of Neurology, \and Neuroimaging Research Core Facility, Medical University of Innsbruck, 6020 Innsbruck, Austria\\
	\email{\{christian.gapp\textsuperscript{\textbf{*}}, elias.tappeiner, martin.welk, rainer.schubert\}@umit-tirol.at,\\ karl.fritscher@vascage.at\\ \{stephanie.mangesius, constantin.eisenschink, philipp.deisl, michael.knoflach, astrid.grams, elke.gizewski\}@i-med.ac.at}
}
%\end{comment}

\begin{comment}
\author{Anonymized Authors}  %% Added for anonymized MICCAI submission
\authorrunning{Anonymized Author et al.}
\institute{Anonymized Affiliations \\
    \email{email@anonymized.com}}
\end{comment}
  
\maketitle              % typeset the header of the contribution
\begin{abstract}
	Multimodal stroke recurrence prediction requires effective integration of heterogeneous clinical and imaging data, yet modality imbalance often causes models to over-rely on dominant modalities and underutilize complementary information. While self-supervised pretraining and selective parameter freezing are commonly employed to improve representation learning and fine-tuning stability, their effect on modality contributions and cross-modal behavior in multimodal medical models remains largely unexplored.
	In this work, we investigate whether image pretraining on 3D CTA scans reduces modality imbalance and improves cross-modal integration for stroke recurrence prediction, a clinically critical task we recently addressed. To this end, two multimodal neural networks are pretrained in a self-supervised manner and subsequently fine-tuned using two distinct freezing strategies. Their performance and modality utilization are compared against both the baseline model from our previous work and models trained entirely from scratch in this study.
	Our results demonstrate that self-supervised pretraining enables more effective utilization of the multimodal image-tabular dataset, outperforming both the prior baseline and all non-pretrained models. Notably, the best-performing Vision Transformer based neural network successfully overcomes unimodal collapse. Synergy analysis reveals significant interactions between vision and both gender and CHD, suggesting clinically relevant patterns for stroke recurrence.
	Overall, our findings demonstrate that self-supervised pretraining and strategic fine-tuning support more balanced modality utilization and enable meaningful cross-modal interactions. Code is publicly available at \githubrepo.
	
	
\end{abstract}
\section{Introduction}
%MAIN PARAGRAPHS:
% Multimodal AI, limitations, problems, solutions

Medical patient data are inherently multimodal, commonly including images (e.g., X-rays, MRI, CT), tabular information, and clinical reports. However, applying multimodal deep learning to medical tasks remains challenging. Despite numerous fusion strategies \cite{MultiModalTransformersSurvey}, models often suffer from unbalanced modality contributions or even modality collapse, where only a subset of modalities is effectively utilized \cite{JavaloyModalityCollapse}. In such cases, unimodal models may outperform multimodal approaches \cite{Wand2020}. Possible causes include conflicting gradients between modalities \cite{JavaloyModalityCollapse} and the strong dependence of fusion strategies on the dataset \cite{Ma2022_Fusion_Strategy_and_dataset}.
Vision Transformer (ViT)-based \cite{16x16WORDS} multimodal architectures have been observed to suffer from unimodal collapse, particularly in vision-language tasks \cite{coffeBean}. However, pretraining can substantially improve downstream performance \cite{RePre2022}, and ViTs often generalize better than ResNets \cite{chen2022_ViTs_vs_ResNets}. While multimodal pretrained models such as CLIP \cite{CLIP} and BiomedCLIP \cite{BiomedCLIP} achieve strong results, they are difficult to apply to many medical image-tabular datasets. Instead, modalities can be pretrained separately. For vision, self-supervised pretraining helps learn meaningful image representations that can subsequently improve fine-tuning performance \cite{tang2022self}.

%This study analyzes the impact of self-supervised image pretraining on the performance and modality contribution of multimodal neural networks. We build upon the image-tabular dataset from \cite{XSRD-Net_Gapp}, where a ResNet \cite{ResNet}-based model was trained for stroke relapse detection using 3D CTAs, tabular patient data, and follow-up information from 119 patients.

This study investigates the effects of self-supervised image pretraining and selective weight freezing during fine-tuning on performance, modality contribution, and cross-modal learning in multimodal stroke recurrence prediction. We build upon the image-tabular dataset from \cite{XSRD-Net_Gapp}, where a ResNet-based \cite{ResNet} model was trained for stroke relapse detection using 3D CTA scans, tabular patient data (age, gender, CHD, PAD), and follow-up information from 119 patients. Importance analysis indicated a balanced use of visual and tabular data, with the arteria carotis communis appearing relevant for relapse prediction.
Here, pretraining enables the use of the full cohort of 491 patients. We pretrain and fine-tune both a ResNet-\cite{ResNet} and a ViT \cite{16x16WORDS}-based multimodal architecture and compare their performance against \cite{XSRD-Net_Gapp} and models trained from scratch in this work. 
Furthermore, we use the modality contribution method from \cite{MCI_Gapp} and the cross-modal feature synergy measure from \cite{Gapp_SyAM} to evaluate the impact of pretraining and strategic fine-tuning on multimodal learning for stroke relapse detection, a largely unexplored topic, within that setup.

\section{Multimodal Stroke Data}
Data acquisition methods and study cohort composition for pretraining and fine-tuning are detailed below, including collection procedures and patient demographics.

\subsection{Data Generation}
As part of the project “Retrospective Pilot Project: Imaging Biomarkers for Vascular Diseases and Vascular Aging”, clinical and imaging data were collected from April 2023 onwards. The cohort included patients with at least one ischemic cerebral event (ICE) who had been admitted to the Stroke Unit of the Department of Neurology, Medical University of Innsbruck, since 2010.
Anonymized imaging data was recorded on Siemens’ syngo.share platform (Version VA32C), while clinical data was collected in a custom database established by an external company.
The anonymized clinical data included patient age, gender, and the occurrence of cardiac or peripheral events. The combined image-tabular dataset was labeled according to the occurrence of recurrent ICE, distinguishing between relapse and non-relapse cases. All CT angiography data from routine diagnostics were fully anonymized prior to analysis.
The study was approved by the local institutional review board (IRB) of the Medical University of Innsbruck (EK-Nr: 1429/2021).

\subsection{Study Population}
After the data cleaning processes, we retained 491 patients with fully usable vision and clinical tabular data.
For the self-supervised pretraining task, 393 of them were used for training, 98 for validation, yielding an approximate 80:20 ratio.
For the fine-tuning task, which involves predicting RFS (relapse free survival) time and classifying patients into relapse and non-relapse groups, the dataset comprises 119 patients. Patients were selected based on their label (relapse vs. relapse-free) and RFS time, as they met the necessary criteria for this task (see \cite{XSRD-Net_Gapp}). Specifically, relapse patients with an RFS below 1,642 days and non-relapse patients with an RFS above 1,825 days were included, with an additional cut-off at 2,555 days applied for non-relapse cases.
The fine-tuning dataset was split into training (95 image-tabular pairs, including 32 relapses) and testing (24 image-tabular pairs, 9 relapses) sets. Thereby it is ensured that the 24 patients used for evaluation were entirely part of the pretraining validation dataset, thus completely excluded from all training steps. A summary of the demographic and clinical characteristics of the study population for the fine-tuning task is provided in \cref{tab:statistical_details_stuy_pop}.

\setlength{\tabcolsep}{10pt}
\begin{table}[t!]
	\centering
	\caption{Baseline characteristics of the study population (N = 119).}
	\label{tab:statistical_details_stuy_pop}
	\begin{tabular}{llcc}
		\toprule
		characteristic & attribute & n & \% \\
		\cmidrule{1-4}
		\textbf{gender} & men  & 79 & 66.4 \\
		& women & 40 & 33.6 \\
		\addlinespace[2mm]
		\textbf{heart disease} & CHD only & 24 & 20.2 \\
		& PAD only & 20 & 16.8 \\
		& CHD + PAD & 6 & 5.0 \\
		& none & 69 & 58.0 \\
		\addlinespace[2mm]
		\textbf{relapse status} & no relapse & 78 & 65.6 \\
		& relapse & 41 & 34.4 \\
		\cmidrule{1-4}
		& & \multicolumn{2}{c}{mean $\pm$ SD} \\
		\cmidrule{3-4}
		\textbf{age (years)} & &\multicolumn{2}{c}{$69.10 \pm 10.18$} \\
		\bottomrule
	\end{tabular}
\end{table}

\subsection{Data Preprocessing}
In the whole study, for both pretraining and fine-tuning tasks, we used 3D CTA scans and tabular data recorded at the time of the first, initial stroke event. The labels (RFS, occurrence of relapse) were finalized at the end of the follow up.
\paragraph{Vision} All images are registered to a fixed, representative image using affine transformations to ensure consistent alignment and comparable properties for downstream deep learning analyses.The images have a size of $224 \times 224 \times 320$~voxels with an isotropic spacing of $1 \times 1 \times 1$~mm.

\paragraph{Tabular}The clinical tabular data includes information about the patients' age, gender, and the heart diseases CHD and PAD. Heart diseases are encoded using a single bit each, while gender is represented with two bits. Age values are z-normalized across the entire dataset.

\section{Training Configurations}
Training is conducted in two stages: first, pretraining of the visual networks, and second, fine-tuning of the multimodal neural networks. We pretrain two visual nets -- ResNet\-Auto\-Enc and Vision\-Auto\-Enc -- and later fine-tune multimodal neural networks utilizing either a ResNet \cite{ResNet} or ViT \cite{16x16WORDS} backbone model. Given the inherently low dimensionality of the tabular data (four attributes), pretraining is unnecessary. Instead, a compact MLP with a single hidden layer is trained from scratch. The core focus of this work lies in advancing the visual modality.

\subsection{Pretraining: Self-Supervised Learning (SSL)}
Motivated from \cite{tang2022self}, we apply a self-supervised learning approach as a pretraining step. To this end, the 3D CTA scans are processed by an autoencoder (comprising an encoder and decoder) that learns to extract features from the visual input and to reconstruct the same input image, optimized with an L1 loss. The encoder part is leveraged for the fine-tuning task. Details to the pretraining architectures are presented below. Additional information is made available at \githubrepo.

\begin{figure}[]
	\centering
	\includegraphics[width=0.9\textwidth]{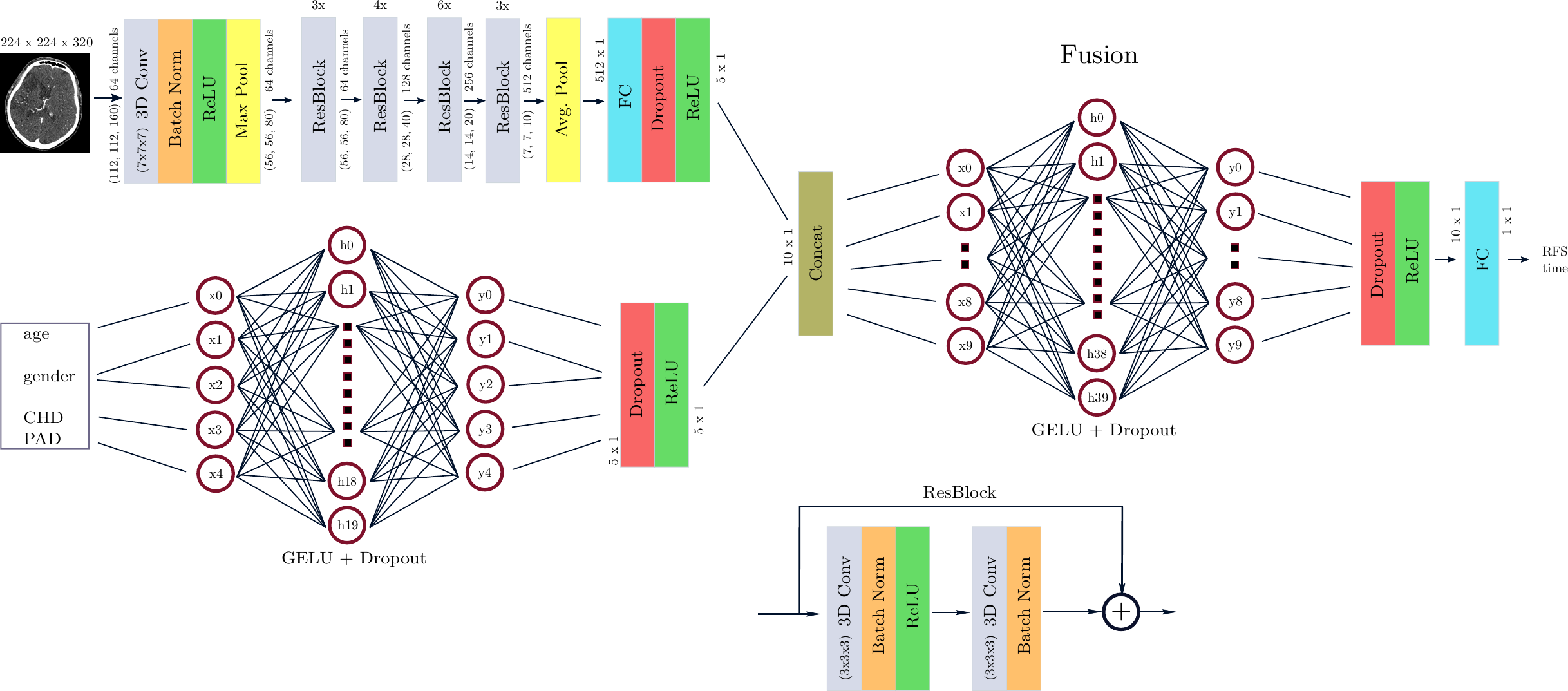}
	\caption{ResNetMLP: Vision model: ResNet34, Tabular model: MLP with one hidden layer. Fusion model: MLP with one hidden layer.}
	\label{fig:ResNetMLP}
	%\end{figure}
	%\begin{figure}[t!]	%removed as figures now stuck together
	%	\centering
	\vspace{1em}
	\includegraphics[width=0.9\textwidth]{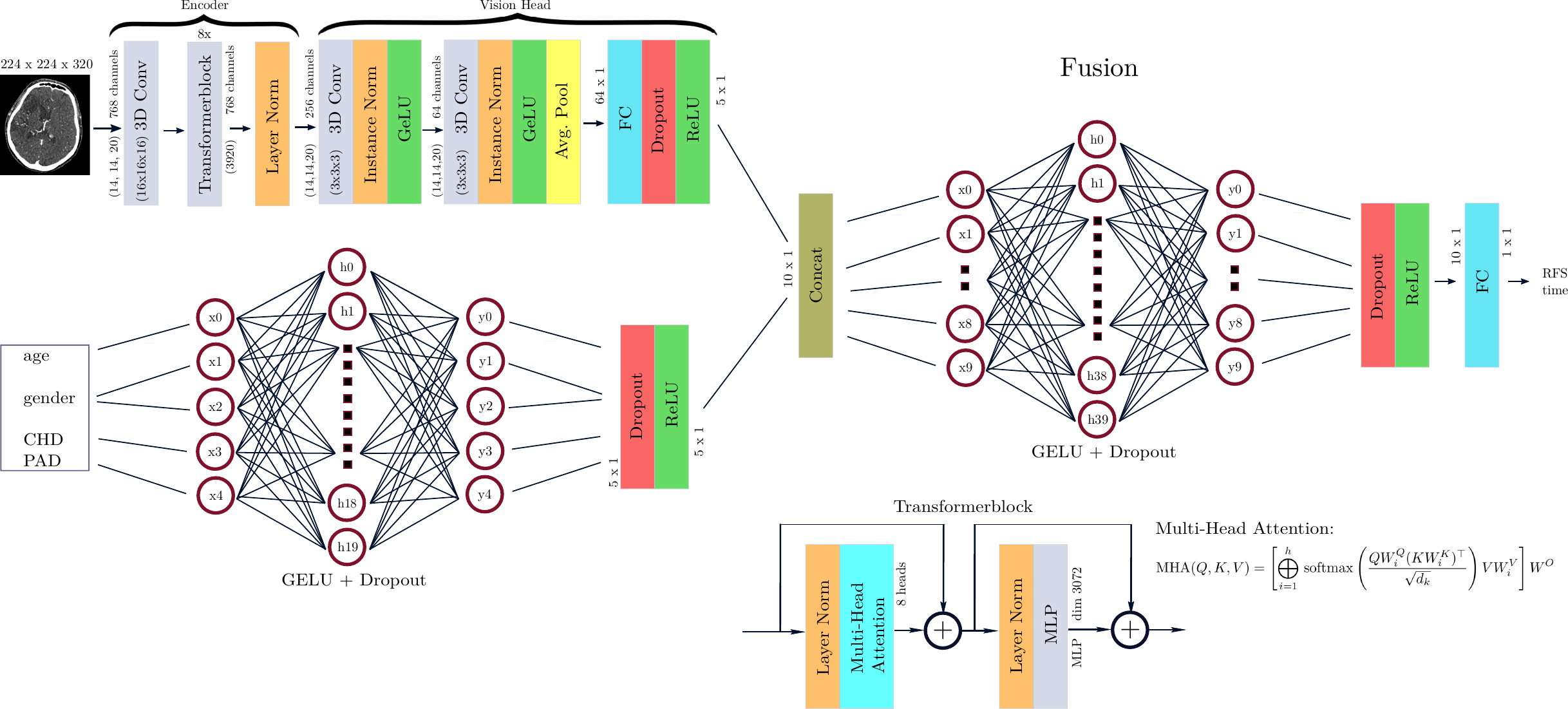}
	\captionof{figure}{ViTMLP: Vision model: ViT with eight heads and eight layers, Tabular model: MLP with one hidden layer, Fusion model: MLP with one hidden layer.}
	\label{fig:ViTMLP}
\end{figure}

\paragraph{ResNetAutoEnc}
The encoder is a ResNet34 (see also vision backbone model in \cref{fig:ResNetMLP}). In the decoder, four convolutional layers without skip connections are applied to the encoded features to reconstruct the image.

\paragraph{ViTAutoEnc}
A Vision Transformer architecture with eight layers and eight attention heads is representing the encoder part (see also backbone model in \cref{fig:ViTMLP}). The decoder, built with two convolutional layers and no skip connections, reconstructs the image.

\paragraph{Computation Time} Pretraining for both models required approximately four days for 300 epochs on an NVIDIA L40S 48GB GPU.

%TODO: Attention: see figure and caption: n hidden layers

\subsection{Fine-tuning Task: Stroke Recurrence Prediction}

As fine-tuning task the stroke relapse prediction is done the same way as in \cite{XSRD-Net_Gapp}. Therein, RFS prediction is performed as a regression task, followed by classification into relapse versus non-relapse groups. Classification is based on the predicted RFS time, using a threshold $\kappa$ in days: predictions below this threshold are considered relapses, while predictions above it are considered non-relapses. Thresholds within the range $\kappa_\mathrm{low} \le \kappa \le \kappa_\mathrm{high}$ were explored in \cite{XSRD-Net_Gapp}. However, in this work we keep $\kappa$ fixed at $\kappa_\mathrm{low}$ = 1,642.
%For this the XSRD-net -- ResNetMLP -- from \cite{XSRD-Net_Gapp} is modified a little. We increased the number of hidden layers in the tabular and fusion MLPs. 
%TODO choose correct option (depending on hidden layer amount)
Using this setting, we fine-tune the XSRD-net architecture proposed in \cite{XSRD-Net_Gapp}, a ResNetMLP, consisting of a ResNet-34 backbone, an MLP backbone, and an MLP classifier. In addition, we fine-tune a ViTMLP employing a ViT (eight heads, eight layers) as vision backbone model. Architectural details are depicted in \cref{fig:ResNetMLP,fig:ViTMLP}.

\paragraph{Model Complexity} The ResNet and ViT backbones have around 63.5M and 59.8M parameters, respectively. As the vision head for ViT has around 5.75M parameters, compared to only 2.56k for the the ResNet34, the vision models are comparable in size. Including tabular MLP 
%(645), %with 2 hidden layers
%TODO: comment correct line
(225), %with one hidden layer
and fusion MLP 
%(5.770T), %with 4 hidden layers
%TODO: comment correct line
(850), %with one hidden layer
the total model sizes amount to roughly 63.60M for ResNetMLP and 65.55M for ViTMLP. Even though the MLPs for the tabular and fusion components are comparatively small, they are still sufficient and play an essential role in the multimodal task. The tabular MLP processes five inputs: one continuous feature for age and four binary features for gender (2 bits) and heart disease (2 bits for CHD and PAD). The fusion MLP combines two $5 \times 1$ outputs from the individual models to a $10 \times 1$ representation and learns joint features prior to the final classification head.

\paragraph{Computation Time} Fine-tuning was performed over 250 epochs, requiring approximately 1~hour and 50~minutes per model on the same hardware.

\subsection{Experiments}
%TODO: choose correct figure depending on num stages applied
%\begin{figure}[]
%	\centering
%	\includegraphics[width=0.7\textwidth]{param_schedule_5_stages.pdf}
%	\caption{Visualization of parameter schedule during progressive layer unfreezing over 250 fine-tuning epochs for ResNet and ViT backbones. The training is divided into five stages, indicated by shaded background regions. Step curves show the cumulative number of trainable parameters (in millions) over time. ViT starts with more trainable parameters due to its larger vision head (see stage~1), whereas ResNet releases more parameters in stages~2-4, as larger layers are affected there.}
%	\label{fig:freezing_strategy}
%\end{figure}
\begin{figure}[t!] %TODO: b! for anonymous. t! for final
	\centering
	\includegraphics[width=0.65\textwidth]{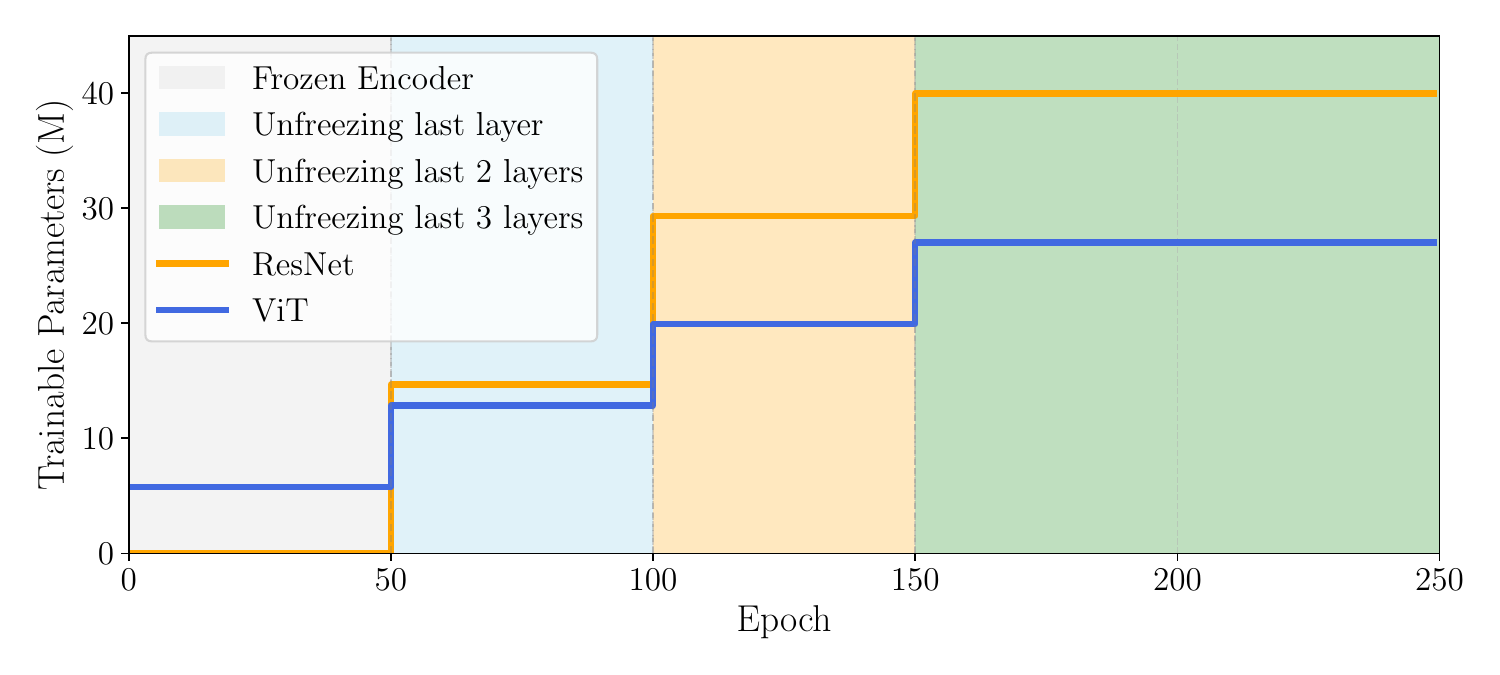}
	\caption{Visualization of parameter schedule during progressive layer unfreezing over 250 fine-tuning epochs for ResNet and ViT backbones. The training is divided into four stages, indicated by shaded background regions. Step curves show the cumulative number of trainable parameters (in millions) over time. ViT starts with more trainable parameters due to its larger vision head (see stage~1), whereas ResNet releases more parameters in stages~2-4, as larger layers are affected there.}
	\label{fig:freezing_strategy}
\end{figure}

To assess the effect of pretraining on the fine-tuning task, we fine-tune pretrained multimodal models for each backbone, ResNet and ViT, and compare them to models trained from scratch -- end-to-end multimodal baseline models.
For this the AUC is computed as performance metric. The modality contribution is measured with the method from \cite{MCI_Gapp}. This occlusion sensitivity based method includes a hyper-parameter $h_\mathrm{modality}$, which defines how many sequences are occluded at a single model forward pass. We use $h_\mathrm{vision}=1$, occluding the entire image at once, and $h_\mathrm{tabular}=4$, where each attribute -- age, gender (two bits occluded at once), CHD, and PAD -- is occluded separately. Synergy analysis is performed using the method introduced in \cite{Gapp_SyAM}, which decodes modality interactions to reveal complementary information at the multimodal feature level.

For each of the two pretrained multimodal models we fine-tune one multimodal model with a fully frozen backbone encoder, and another one following the freezing strategy illustrated in \cref{fig:freezing_strategy}, which successively unfreezes later backbone layers after 50, 100 and 150 epochs.
A large portion of the (pretrained) encoder weights is thus frozen throughout the entire 250 epochs. Weights from the last layer are trained 200, weights from the second last layer 150 and weights from the third last layer 100 epochs. Both strategies of either fully freezing or partly freezing encoder weights, rather than to just fine-tune the whole pretrained model without freezing, are used to mitigate the risk of catastrophic forgetting, a well-known challenge in sequential learning \cite{French1999,GoodfellowcatastrophicForgetting}.
By applying freezing approaches, previously learned features are preserved, allowing the model to effectively adapt to new data or, in our case, new tasks.
All in all, thus six multimodal neural networks (for each ViT, ResNet backbone: two pretrained + one from scratch) are trained.

\section{Results}

%TODO: check if correct curves plotted
\begin{figure}[] %TODO: b! for anonymous, [] for final
	\centering
	\begin{minipage}{0.48\textwidth}
		\centering
		\includegraphics[width=\textwidth]{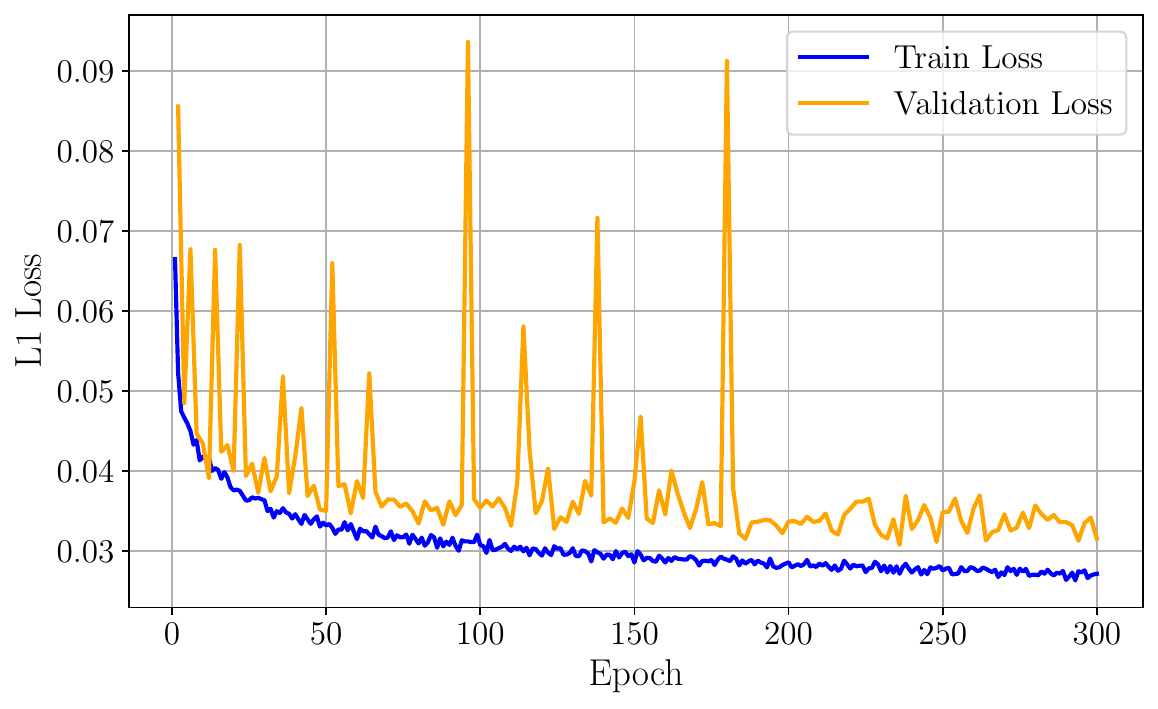}
		\caption*{(a) ResNetAutoEnc}
	\end{minipage}
	\hfill
	\begin{minipage}{0.48\textwidth}
		\centering
		\includegraphics[width=\textwidth]{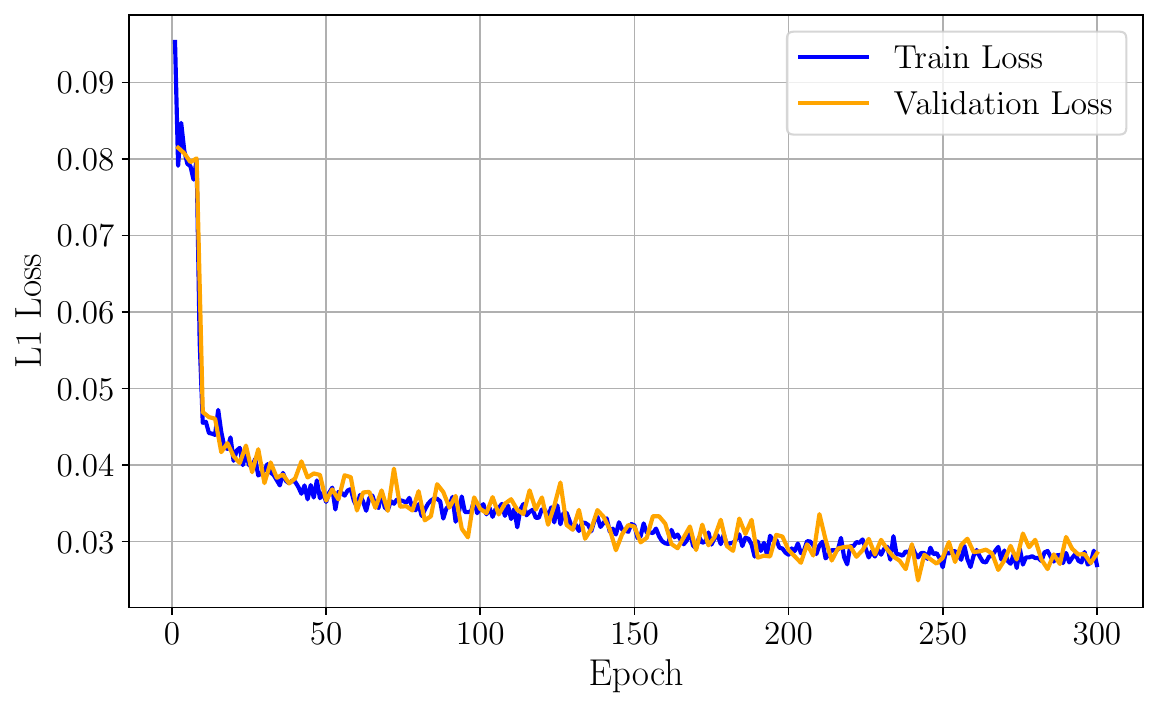}
		\caption*{(b) ViTAutoEnc}
	\end{minipage}
	\caption{Pretraining L1 loss curves over epochs for training and validation across both architectures.}
	\label{fig:pretrain_resnet_vit}
\end{figure}

\paragraph{Pretraining}
In \cref{fig:pretrain_resnet_vit}, we plot the pretraining L1-losses of the two autoencoder backbones: ResNetAutoEnc and ViTAutoEnc.

\paragraph{Fine-tuning Task}
\cref{tab:performance_results,tab:mci_results_classification} summarize the effects of pretraining on multimodal disease classification performance and modality contributions. \cref{tab:performance_results} compares the overall AUC and the relative impact of the vision modality ($mc_v$) between ResNetMLP and ViTMLP architectures under different training setups. \cref{tab:mci_results_classification} provides a more detailed breakdown of individual modality contributions \cite{MCI_Gapp}, highlighting how vision and tabular features (age, gender, CHD, PAD) are utilized by the model when using pretrained weights compared to training from scratch. \cref{tab:synergy} details the synergy-related analysis results \cite{Gapp_SyAM} for the best-performing models of each architecture. Additionally, an analysis of a single relapse case is provided in \cref{app-sec:Interpretability_Analysis}, as shown in \cref{fig:vision_interpretability}.

\setlength{\tabcolsep}{4pt}
\begin{table}[]
	\centering
	\caption[ResNetMLP vs. ViTMLP: AUC and vision modality contribution by training type.]{Pretraining effects on performance for ResNetMLP and ViTMLP. The last column shows the modality contribution of vision $mc_v$. The whole analysis is done with the test dataset. Highlighted in green: overall best model.}
	\label{tab:performance_results}
	\begin{tabular}{llccc}
		\toprule
		XSRD-Net & Training Setup & train AUC & test AUC & $mc_v$ \\
		\midrule
		{ResNetMLP} & from scratch -- \emph{baseline} &$0.570$ & $0.549$ & 79.46\% \\ %run_43
		& pretrained frozen &$0.409$ & $0.653$ & 93.31\% \\ %run_45
		& \textbf{pretrained temp. frozen} &$0.695$ & $\mathbf{0.715}$ & 92.20\% \\ %run_42
		\addlinespace
		{ViTMLP} & from scratch -- \emph{baseline} &$0.617$ & $0.722$ & \hphantom{0}0.13\% \\ %run_44
		&\cellcolor{softgreen}\textbf{pretrained frozen} &\cellcolor{softgreen}$0.805$ & \cellcolor{softgreen}$\mathbf{0.743}$ & \cellcolor{softgreen}28.39\% \\ %run_46
		& pretrained temp. frozen &$0.626$ & $0.729$ & \hphantom{0}5.53\% \\ %run_41
		\bottomrule
	\end{tabular}
\end{table}

%\vspace{2em}

\setlength{\tabcolsep}{4pt}
\begin{table}[]
	\centering
	\caption{Detailed modality contributions for RFS predictions using the test dataset. Green highlighted: modality contributions for overall best model from \cref{tab:performance_results}. Marked in red: unimodal collapse of ViTMLP trained from scratch.}
	\label{tab:mci_results_classification}
	\begin{tabular}{llccccc}
		\toprule
		XSRD-Net & Training Setup & vision & \multicolumn{4}{c}{tabular}\\
		%\cmidrule(lr){1-1}
		%\cmidrule(lr){2-2}
		%\cmidrule(lr){3-6}		
		\midrule
		&& 3D CTA & age & gender & CHD & PAD \\
		\cmidrule(lr){3-3}
		\cmidrule(lr){4-4}
		\cmidrule(lr){5-5}
		\cmidrule(lr){6-6}
		\cmidrule(lr){7-7}
		ResNetMLP & from scratch -- \emph{baseline} & 79.46\% & 8.09\% &6.96\% &0.68\% & 4.81\%\\
		& pretrained frozen &93.31\% &2.21\% &3.13\% &0.31\% &1.04\% \\
		& pretrained temp. frozen & 92.20\% &2.57\% &3.91\% &0.40\% &0.92\%\\
		\addlinespace[2mm]
		ViTMLP & from scratch -- \emph{baseline} & \cellcolor{softred}\hphantom{0}0.13\% &12.95\% &75.68\% &4.59\% &6.65\%\\
		& \cellcolor{softgreen}pretrained frozen & \cellcolor{softgreen}28.39\% &\cellcolor{softgreen}8.73\% &\cellcolor{softgreen}55.60\% &\cellcolor{softgreen}3.03\% &\cellcolor{softgreen}4.25\%\\
		& pretrained temp. frozen &5.53\% &11.97\% &71.84\% &4.37\% &6.29\% \\
		\bottomrule
	\end{tabular}
\end{table}

%\vspace{2em}

%results from run_46 (ViTMLP pretrained frozen)
% and from run_42 (ResNetMLP pretrained temp frozen)
\setlength{\tabcolsep}{6pt}
\begin{table}[]
	\centering
	\small
	\caption[Modality contribution \cite{MCI_Gapp} and synergy scores \cite{Gapp_SyAM} for RFS prediction]{Modality contribution (MC) \cite{MCI_Gapp} and synergy scores \cite{Gapp_SyAM} for RFS prediction across 3D CTA ($i$) and individual tabular features ($j$) using the best models for each architecture on the test set. $s_{ij \mathrm{(relapses)}}$ is computed on relapse cases only. Green highlighted: significant synergy.}
	\label{tab:synergy}
	\begin{tabular}{l c c c c c}
		\toprule
		\multicolumn{6}{l}{ViTMLP model (pretrained frozen)}\\
		\midrule
		& {3D CTA} & {age} & {gender} & {CHD} & {PAD} \\
		\midrule
		{MC}
		& 0.284
		& 0.087
		& 0.556
		& 0.030
		& 0.043\\
		{$s_{ij}$} 
		& $\times$
		& $-0.148$
		& $+0.337$
		& $+0.105$ 
		& $+0.002$\\
		{$s_{ij \mathrm{(relapses)}}$}
		& $\times$
		& $-0.436$
		&\cellcolor{softgreen}$+1.068$
		& \cellcolor{softgreen}$+0.300$ 
		& $+0.004$\\
		%{$s_{ij} \mathrm{(non-relapses)}$}
		\bottomrule
		%\end{tabular}
		&&&&&\\
		%results from run_42 (ResNetMLP pretrained temp. frozen)
		%	\begin{tabular}{l c c c c c}
			\toprule
			\multicolumn{6}{l}{ResNetMLP model (pretrained temp. frozen)}\\
			\midrule
			& {3D CTA} & {age} & {gender} & {CHD} & {PAD} \\
			\midrule
			{MC}
			& 0.922
			& 0.026
			& 0.039
			& 0.004
			& 0.009\\
			{$s_{ij}$} 
			& $\times$
			& $-0.008$
			& $0.022$
			& $0.002$ 
			& $-0.013$\\
			{$s_{ij \mathrm{(relapses)}}$}
			& $\times$
			& $-0.015$
			& $0.026$
			& $0.006$ 
			& $-0.013$\\
			%{$s_{ij} \mathrm{(non-relapses)}$}
			\bottomrule
		\end{tabular}
	\end{table}
	
	\paragraph{Significance Tests for ViTMLP}
	Statistical power analysis indicates that the pretrained, frozen ViTMLP significantly outperforms the ViTMLP trained from scratch in terms of AUC,
	%on the training/testing dataset
	with a bootstrap p-value of 0.003 (< 0.05 significance level). Similarly, compared to the ResNetMLP trained from scratch, the bootstrap p-value is 0.0026, also indicating a statistically significant difference. In contrast, no statistically significant difference was observed between the ResNetMLP models (trained from scratch versus pretrained with temporary freezing), as indicated by a bootstrap p-value of 0.086 (> 0.05 level).

	\section{Discussion}
	
	%new:
	%1 SSL general + results
	%2 Fine-tuning results comparing against baselines
	%3 Fine-tuning each (integrated paragraph with the one above)
	%4 MCI + synergy
	
	%1
	% SSL for full medical dataset usage
	Self-supervised visual pretraining enables the use of all patient data in a cohort, independent of disease-specific labels or exclusion criteria.
	% compuation time, efficient fine-tuning
	Despite full cohort utilization, computation is efficient: pretraining takes a few days, while fine-tuning requires less than two hours. Moreover, models can be efficiently updated with newly collected data.
	% Results for pretraining
	Results from pretraining (\cref{fig:pretrain_resnet_vit}) show that the ViTAutoEnc reconstructs images more effectively than the ResNetAutoEnc, with similar train and validation L1 losses, while the ResNetAutoEnc shows slight overfitting. These differences are also reflected in fine-tuning performance.
	
	%2
	% Fine-tuning Task
	% comparing to XSRD
	The ResNetMLP with pretrained ResNet weights and the freezing strategy in \cref{fig:freezing_strategy} achieves an AUC of 0.72 on the test set. The best overall result is obtained by the pretrained ViTMLP with a fully frozen encoder, reaching an AUC of 0.74 (\cref{tab:performance_results}). Both architecture-specific best models improve upon the stroke relapse detection performance reported in \cite{XSRD-Net_Gapp} (AUC: 0.71).
	%
	% comparing each other
	Furthermore, the two best models outperform the baseline models trained from scratch (significant for ViTMLP with p < 0.05). For the ResNetMLP both the model trained from scratch and the pretrained model with fully frozen encoder could not solve the fine-tuning task sufficiently. The ViTMLP model trained from scratch suffered a unimodal collapse, relying only on the tabular data while ignoring the vision input. Although test AUC is reasonably good, the model is underfit (train AUC 0.62), indicating insufficient task learning. The separation between relapse and non-relapse is largely driven by gender and age (see also \cref{tab:mci_results_classification}), which is of limited medical relevance.
	%
	%3
	% ResNetMLP
	The poor performance of the pretrained ResNetMLP with a frozen encoder (AUC 0.65 test, 0.41 train) may be due to limited pretraining and the very small number of trainable parameters in the vision head (only a classification layer). This explains why the progressive unfreezing strategy is effective: successively unfreezing larger layers allows the model to learn more relevant patterns in later epochs.
	%
	% ViTMLP
	In contrast, the ViTMLP shows the opposite behaviour. Due to strong pretraining, the best performance is achieved with a pretrained, fully frozen encoder and a learnable vision head comprising convolutional layers and a classifier (see \cref{tab:performance_results}). Applying successive layer unfreezing reduces performance, especially training AUC, and likely generalizability, as the encoder adapts to the fine-tuning task at the expense of pretrained representations.
	
	%4
	% MCI
	Modality contributions measured with \cite{MCI_Gapp} highlight the effect of pretraining for multimodal integration. The ResNetMLP uses mainly visual data (79.5\%), increasing to 92.20\% (partly frozen) and 93.31\% (frozen encoder) in pretrained variants. (\cref{tab:mci_results_classification}). In contrast, pretraining the ViTMLP prevents unimodal collapse, raising vision contribution from 0.13\% to 5.53\% (partly frozen) and 28.39\% (frozen encoder).
	%
	% Discussion: "ResNetMLP vs ViTMLP mci vision"
	Despite being lower than the ResNetMLPs, the pretrained ViTMLP (frozen encoder) achieves more balanced modality use and better performance by enabling stronger fusion learning.
	%Synergy
	Synergy analysis confirms meaningful cross-modal interactions, with positive synergy between gender and vision (+0.337 / +1.068) and CHD and vision (+0.105 / +0.300), while age shows redundancy and PAD near independence (\cref{tab:synergy}). The ResNetMLP exhibits weaker interaction effects due to its stronger reliance on vision.
	
	\section{Conclusion}
	
	Through utilization of self-supervised pretraining with 3D CTA scans we could improve results from \cite{XSRD-Net_Gapp} for early detection of stroke relapses. While the pretrained ResNetMLP achieved its best performance after fine-tuning with progressive layer unfreezing, the pretrained ViTMLP with a frozen encoder yielded the best overall performance, with an AUC of 0.74 on the test set. The contribution of the visual modality increased substantially throughout the pretraining stage across all models, provided that freezing was applied judiciously to specific parts of the network. Especially for ViTMLP, we could even tackle an unimodal collapse that occurred for the baseline model trained from scratch. Vision features were used considerably more effective in the multimodal task (0.13\% before vs. 28.39\% after pretraining). Hence, pretraining unlocked a broader range of discriminative features for the stroke relapse detection task.
	By enhancing the contribution of individual modalities, multimodal datasets can be fully leveraged to reveal vision patterns linked to attributes such as gender, or cardiovascular disease, as synergy related analysis suggested.
	%Outlook
	%
	With our contribution to the field of multimodal, medical, explainable AI, we aim to inspire future research in this area to use pretraining methods to reveal otherwise hidden information embedded in the data. Self-supervised pretraining can definitely lead to improvements across multiple fine-tuning tasks. 
	While this involves leveraging multimodal features, to the best of our knowledge, we are the first to demonstrate the impact of both image pretraining and strategic multimodal fine-tuning on modality contributions and cross-modal synergy.
	%Specifically, we analyze the modality contribution gap between pretrained models and those trained from scratch, revealing novel insights at the modality level.
	%Code
	%Removed:
	%Due to legal restrictions, the dataset cannot be publicly released.
	The full pretrain code, partial downstream-task components, and supplementary materials are available at \githubrepo.
	
\begin{credits}
		\subsubsection{\ackname}
		This study is partly supported by VASCage -- Centre on Clinical Stroke Research.
				
		\subsubsection{\discintname}
		The authors have no competing interests to declare that are
		relevant to the content of this article.

\end{credits}
	
		\bibliographystyle{splncs04}
		\bibliography{references}

@Article{MCI_Gapp,
	author={Gapp, Christian
	and Tappeiner, Elias
	and Welk, Martin
	and Fritscher, Karl
	and Gizewski, Elke R.
	and Schubert, Rainer},
	title={What are you looking at? Modality contribution in multimodal medical deep learning},
	journal={International Journal of Computer Assisted Radiology and Surgery},
	year={2025},
	month={Oct},
	day={02},
	issn={1861-6429},
	doi={10.1007/s11548-025-03523-w},
}

@misc{XSRD-Net_Gapp,
      title={XSRD-Net: EXplainable Stroke Relapse Detection}, 
      author={Christian Gapp and Elias Tappeiner and Martin Welk and Karl Fritscher and Stephanie Mangesius and Constantin Eisenschink and Philipp Deisl and Michael Knoflach and Astrid E. Grams and Elke R. Gizewski and Rainer Schubert},
      year={2025},
      eprint={2509.07772},
      archivePrefix={arXiv},
      primaryClass={cs.CV},
      url={https://arxiv.org/abs/2509.07772}, 
}

@article{Gapp_SyAM,
  author       = {Gapp, Christian and Tappeiner, Elias and Welk, Martin and Fritscher, Karl and Mangesius, Stephanie and Eisenschink, Constantin and Deisl, Philipp and Knoflach, Michael and Grams, Astrid E. and Gizewski, Elke R. and Schubert, Rainer},
  title        = {Decoding Modality Interactions in Medical AI With {SyAM}: A Light-weight Synergy-Augmented Metric},
  year         = {2026},
  month        = {03},
  note         = {Accepted at CARS 2026 (to appear); preprint available at ResearchGate},
  doi          = {10.13140/RG.2.2.28330.58566},
}

@inproceedings{tang2022self, 
	title={Self-supervised pre-training of swin transformers for 3d medical image analysis}, 
	author={Tang, Yucheng and Yang, Dong and Li, Wenqi and Roth, Holger R and Landman, Bennett and Xu, Daguang and Nath, Vishwesh and Hatamizadeh, Ali}, 
	booktitle={Proceedings of the IEEE/CVF Conference on Computer Vision and Pattern Recognition}, 
	pages={20730--20740}, 
	year={2022} 
}

@inproceedings{16x16WORDS,
	title={An Image is Worth 16x16 Words: Transformers for Image Recognition at Scale},
	author={Alexey Dosovitskiy and Lucas Beyer and Alexander Kolesnikov and Dirk Weissenborn and Xiaohua Zhai and Thomas Unterthiner and Mostafa Dehghani and Matthias Minderer and Georg Heigold and Sylvain Gelly and Jakob Uszkoreit and Neil Houlsby},
	booktitle={International Conference on Learning Representations},
	year={2021},
	eprint={2010.11929},
	archivePrefix={arXiv},
	primaryClass={cs.CV},
	doi = {10.48550/arxiv.2010.11929}
}

@INPROCEEDINGS{ResNet,
	author={He, Kaiming and Zhang, Xiangyu and Ren, Shaoqing and Sun, Jian},
	booktitle={2016 IEEE Conference on Computer Vision and Pattern Recognition (CVPR)}, 
	title={Deep Residual Learning for Image Recognition}, 
	year={2016},
	month={06},
	volume={},
	number={},
	pages={770-778},
	doi = {10.1109/CVPR.2016.90}
}

@misc{GoodfellowcatastrophicForgetting,
      title={An Empirical Investigation of Catastrophic Forgetting in Gradient-Based Neural Networks}, 
      author={Ian J. Goodfellow and Mehdi Mirza and Da Xiao and Aaron Courville and Yoshua Bengio},
      year={2015},
      eprint={1312.6211},
      archivePrefix={arXiv},
      primaryClass={stat.ML},
      url={https://arxiv.org/abs/1312.6211}, 
}

@Article{French1999,
	author={French, Robert M.},
	title={Catastrophic forgetting in connectionist networks},
	journal={Trends in Cognitive Sciences},
	year={1999},
	month={Apr},
	day={01},
	publisher={Elsevier},
	volume={3},
	number={4},
	pages={128-135},
	issn={1364-6613},
	doi={10.1016/S1364-6613(99)01294-2},
}

@misc{BiomedCLIP,
      title={BiomedCLIP: a multimodal biomedical foundation model pretrained from fifteen million scientific image-text pairs}, 
      author={Sheng Zhang and Yanbo Xu and Naoto Usuyama and Hanwen Xu and Jaspreet Bagga and Robert Tinn and Sam Preston and Rajesh Rao and Mu Wei and Naveen Valluri and Cliff Wong and Andrea Tupini and Yu Wang and Matt Mazzola and Swadheen Shukla and Lars Liden and Jianfeng Gao and Angela Crabtree and Brian Piening and Carlo Bifulco and Matthew P. Lungren and Tristan Naumann and Sheng Wang and Hoifung Poon},
      year={2025},
      eprint={2303.00915},
      archivePrefix={arXiv},
      primaryClass={cs.CV},
      url={https://arxiv.org/abs/2303.00915}, 
}

@InProceedings{CLIP,
  title = 	 {Learning Transferable Visual Models From Natural Language Supervision},
  author =       {Radford, Alec and Kim, Jong Wook and Hallacy, Chris and Ramesh, Aditya and Goh, Gabriel and Agarwal, Sandhini and Sastry, Girish and Askell, Amanda and Mishkin, Pamela and Clark, Jack and Krueger, Gretchen and Sutskever, Ilya},
  booktitle = 	 {Proceedings of the 38th International Conference on Machine Learning},
  pages = 	 {8748--8763},
  year = 	 {2021},
  editor = 	 {Meila, Marina and Zhang, Tong},
  volume = 	 {139},
  series = 	 {Proceedings of Machine Learning Research},
  month = 	 {18--24 Jul},
  publisher =    {PMLR},
  url = 	 {https://proceedings.mlr.press/v139/radford21a.html},
}

@InProceedings{JavaloyModalityCollapse,
  title = 	 {Mitigating Modality Collapse in Multimodal {VAE}s via Impartial Optimization},
  author =       {Javaloy, Adrian and Meghdadi, Maryam and Valera, Isabel},
  booktitle = 	 {Proceedings of the 39th International Conference on Machine Learning},
  pages = 	 {9938--9964},
  year = 	 {2022},
  editor = 	 {Chaudhuri, Kamalika and Jegelka, Stefanie and Song, Le and Szepesvari, Csaba and Niu, Gang and Sabato, Sivan},
  volume = 	 {162},
  series = 	 {Proceedings of Machine Learning Research},
  month = 	 {17--23 Jul},
  publisher =    {PMLR},
  url = 	 {https://proceedings.mlr.press/v162/javaloy22a.html},
}

@inproceedings{coffeBean,
   title={{MM-SHAP}: A Performance-agnostic Metric for Measuring Multimodal Contributions in Vision and Language Models \& Tasks},
   booktitle={Proceedings of the 61st Annual Meeting of the Association for Computational Linguistics (Volume 1: Long Papers)},
   publisher={Association for Computational Linguistics},
   author={Parcalabescu, Letitia and Frank, Anette},
   DOI={10.18653/v1/2023.acl-long.223},
   year={2023} 
}

@INPROCEEDINGS{Wand2020,
  author={Wang, Weiyao and Tran, Du and Feiszli, Matt},
  booktitle={2020 IEEE/CVF Conference on Computer Vision and Pattern Recognition (CVPR)}, 
  title={What Makes Training Multi-Modal Classification Networks Hard?}, 
  year={2020},
  volume={},
  number={},
  pages={12692-12702},
  doi={10.1109/CVPR42600.2020.01271}}

@INPROCEEDINGS{Ma2022_Fusion_Strategy_and_dataset,
  author={Ma, Mengmeng and Ren, Jian and Zhao, Long and Testuggine, Davide and Peng, Xi},
  booktitle={2022 IEEE/CVF Conference on Computer Vision and Pattern Recognition (CVPR)}, 
  title={Are Multimodal Transformers Robust to Missing Modality?}, 
  year={2022},
  volume={},
  number={},
  pages={18156-18165},
  doi={10.1109/CVPR52688.2022.01764}
}

@ARTICLE{MultiModalTransformersSurvey,
	author={Xu, Peng and Zhu, Xiatian and Clifton, David A.},
	journal={{IEEE} Transactions on Pattern Analysis \& Machine Intelligence},
	title={Multimodal Learning With Transformers: A Survey},
	year={2023},
	volume={45},
	number={10},
	ISSN={1939-3539},
	pages={12113-12132},
	doi={10.1109/TPAMI.2023.3275156},
	publisher={IEEE Computer Society},
	address={Los Alamitos, CA, USA},
	month={10}
}

@inproceedings{RePre2022,
  title     = {RePre: Improving Self-Supervised Vision Transformer with Reconstructive Pre-training},
  author    = {Wang, Luya and Liang, Feng and Li, Yangguang and Zhang, Honggang and Ouyang, Wanli and Shao, Jing},
  booktitle = {Proceedings of the Thirty-First International Joint Conference on
               Artificial Intelligence, {IJCAI-22}},
  publisher = {International Joint Conferences on Artificial Intelligence Organization},
  editor    = {Lud De Raedt},
  pages     = {1437--1443},
  year      = {2022},
  month     = {7},
  doi       = {10.24963/ijcai.2022/200},
}

@inproceedings{chen2022_ViTs_vs_ResNets,
title={When Vision Transformers Outperform ResNets without Pre-training or Strong Data Augmentations},
author={Xiangning Chen and Cho-Jui Hsieh and Boqing Gong},
booktitle={International Conference on Learning Representations},
year={2022},
url={https://openreview.net/forum?id=LtKcMgGOeLt}
}
		
		%APPENDIX
	
\clearpage
		
		\appendix

		\section{Interpretability Analysis for one Recurrence Case} \label{app-sec:Interpretability_Analysis}
%		For the best-performing ReNetMLP and ViTMLP, we analyze one true positive relapse case (TP\textsubscript{1}). \cref{fig:vision_interpretability} shows occlusion maps with a mask size of $[8,8,10]$ (25,088 forward passes).
%		%
%		The ResNetMLP highlights the left neck, consistent with \cite{XSRD-Net_Gapp}, where carotid arteries were relevant for RFS prediction, appearing as a thin spike along the arteria carotis communis (see (a)). The ViTMLP instead focuses on an upper-central, slightly right intracranial region (see (b)).
%		%
%		However, importance differs in magnitude: vision dominates in the ResNetMLP, while gender is more influential in the ViTMLP (35.94\% vs. 8.50\%). This suggests a gender-related visual feature in the ViTMLP.
		
%		We analyze one true positive relapse case (TP\textsubscript{1}) using occlusion maps ([8,8,10], 25,088 passes) to interpret model behavior. ResNetMLP primarily attends to the left neck region around the carotid artery, while ViTMLP shifts focus to an upper-central, slightly right intracranial area. Attribution differs: vision dominates in ResNetMLP, whereas gender is more influential in ViTMLP (35.94\% vs. 8.50\%), suggesting stronger reliance on gender-linked visual cues.
		
		%Removed this table
		% \cref{tab:mci_item_appendix} shows the detailed modality contributions for the item using the two networks.
		
		%\vspace*{\fill}
		%run_53: ResNetMLP pretrained progressive unfreezing
		%run_54: ViTMLP pretrained frozen
		\begin{figure}[]
    \begin{minipage}{0.48\textwidth}
			\centering
			\includegraphics[width=0.245\textwidth]{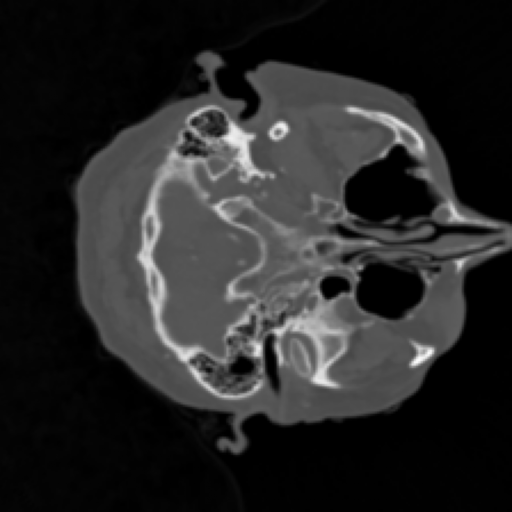} %slice_xy_20 CTA
			\includegraphics[width=0.35\textwidth]{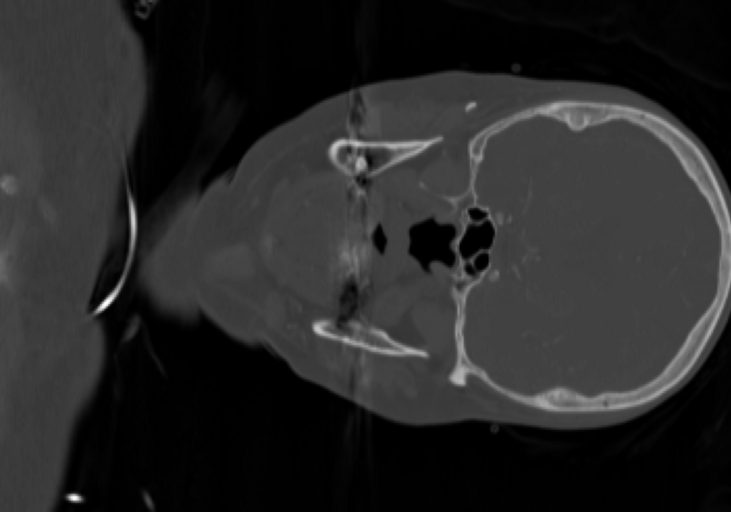} %slice_xz_17 CTA
			\includegraphics[width=0.35\textwidth]{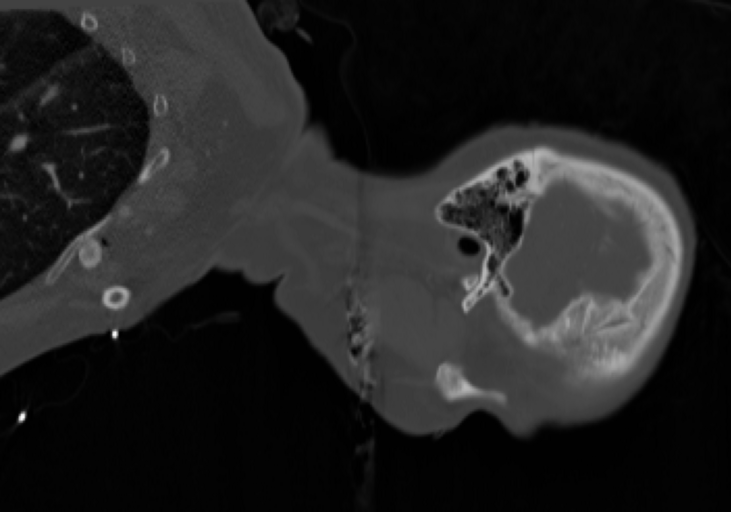} %slice_yz_6 CTA
			\\\vspace{1mm}
			\includegraphics[width=0.245\textwidth]{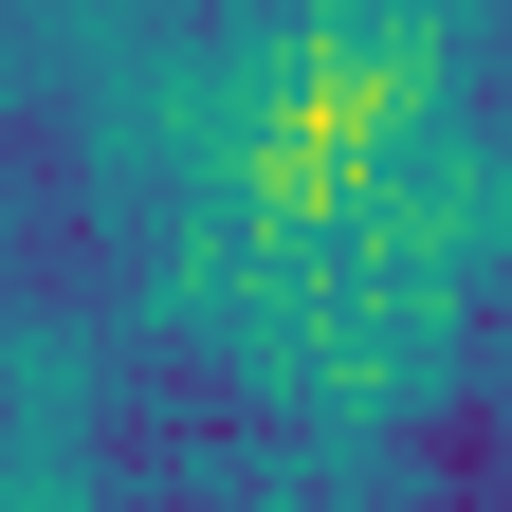} %slice_xy_20 occ sens
			\includegraphics[width=0.35\textwidth]{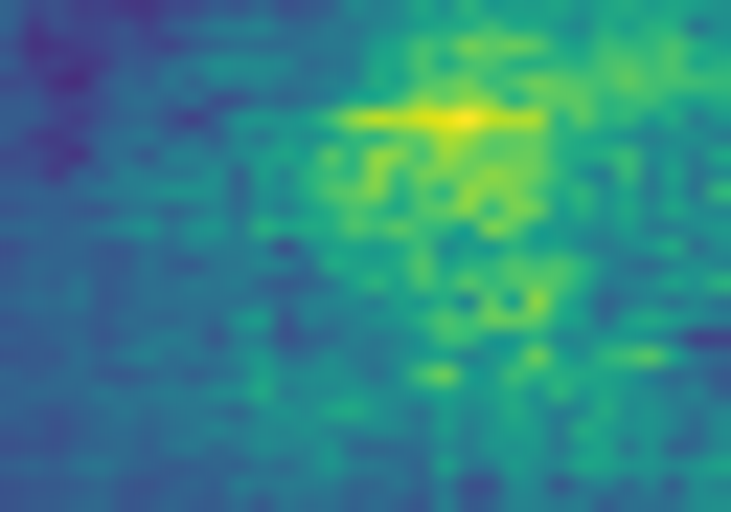} %slice_xz_17 occ sens
			\includegraphics[width=0.35\textwidth]{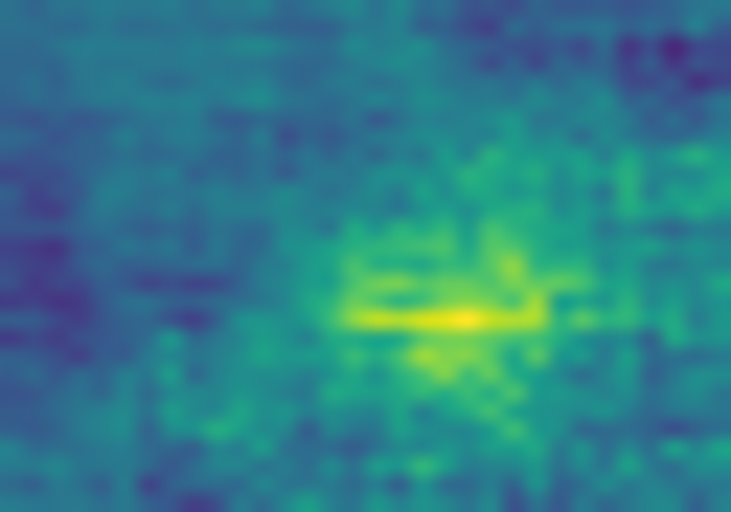} %slice_yz_6 occ sens
			\caption*{(a) Occ.Sens with best ResNetMLP}
    \end{minipage}
		\hfill
    \begin{minipage}{0.48\textwidth}
			\centering
			\includegraphics[width=0.245\textwidth]{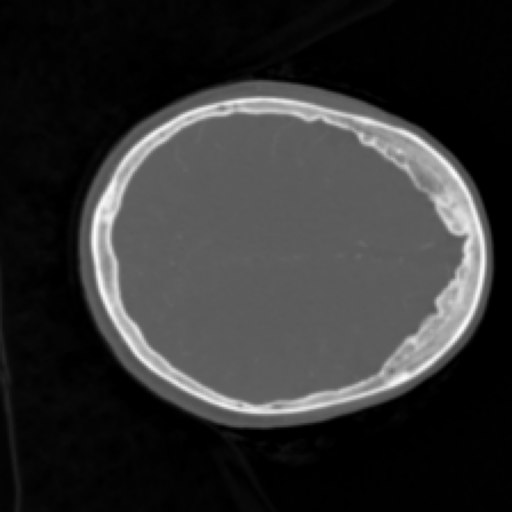} %slice_xy_28 CTA
			\includegraphics[width=0.35\textwidth]{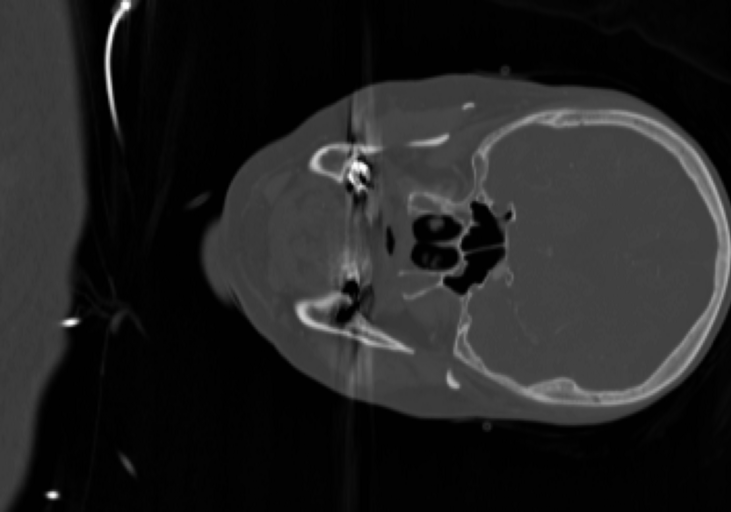} %slice_xz_18 CTA
			\includegraphics[width=0.35\textwidth]{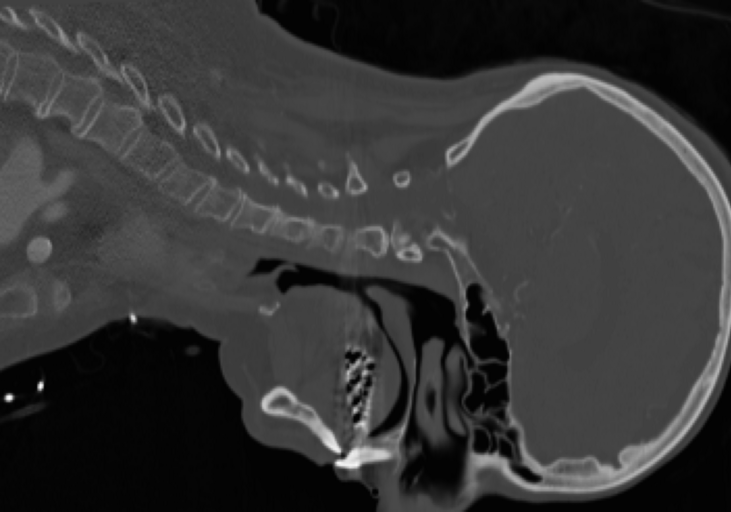} %slice_yz_12 CTA
			\\\vspace{1mm}
			\includegraphics[width=0.245\textwidth]{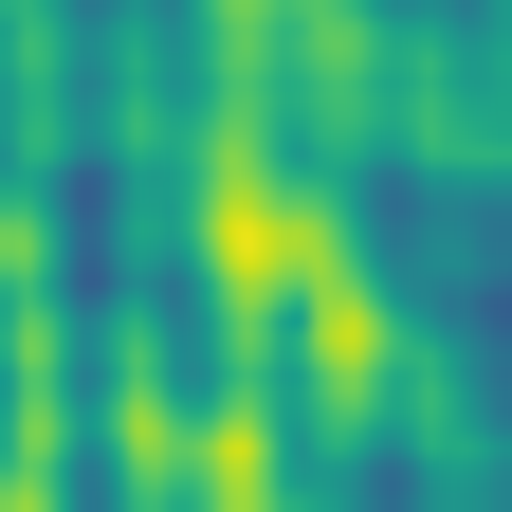} %slice_xy_28 occ sens
			\includegraphics[width=0.35\textwidth]{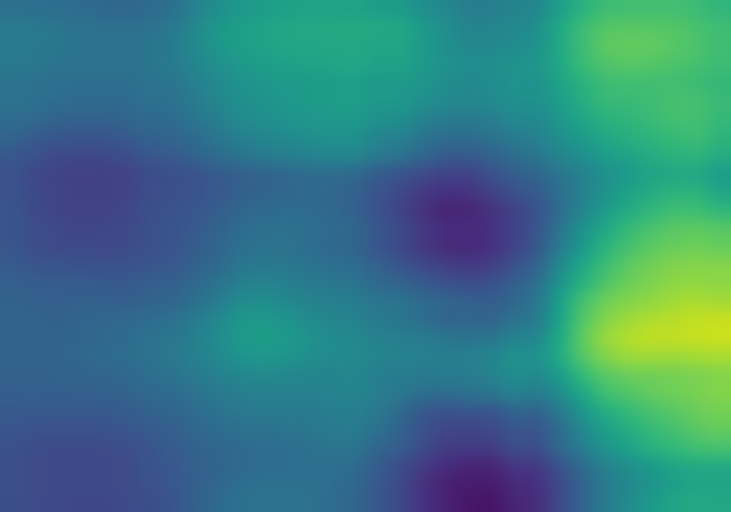} %slice_xz_18 occ sens
			\includegraphics[width=0.35\textwidth]{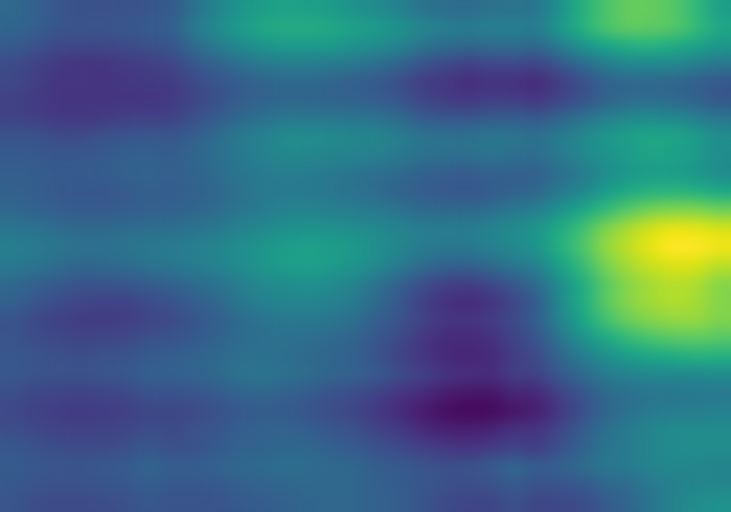} %slice_yz_12 occ sens
			\caption*{(b) Occ.Sens with best ViTMLP}
    \end{minipage}
	\caption{
	Occlusion sensitivity on 3D CTA scans for a true positive relapse example TP\textsubscript{1} (RFS = 107 days), using a mask size of [8,8,10] (25,088 inference passes). (a) ResNetMLP (pretrained, temp. frozen; visual importance 82.40\%), (b) ViTMLP (pretrained, frozen; visual importance 54.28\%): original images (top) and saliency maps (bottom). Slices are transversal (top view), coronal (back view), and sagittal (bottom view); yellow indicates high importance. ResNetMLP focuses on the left carotid region, while ViTMLP shifts to an upper-central intracranial area. Vision dominates in ResNetMLP, whereas gender is more influential in ViTMLP (35.94\% vs. 8.50\%), suggesting stronger reliance on gender-linked visual cues.
	}
		
		\label{fig:vision_interpretability}
\end{figure}
		%\vspace*{\fill}
		
%		\setlength{\tabcolsep}{4pt}
%		\begin{table}[]
%			\centering
%			\caption{Comparison of modality-level contributions for the true positive relapse example TP\textsubscript{1} (RFS = 107 days), evaluated using models trained from scratch and the best-performing ResNetMLP and ViTMLP architectures.}
%			\label{tab:mci_item_appendix}
%			\begin{tabular}{llccccc}
%				\toprule
%				XSRD-Net & Training Setup & vision & \multicolumn{4}{c}{tabular}\\
%				%\cmidrule(lr){1-1}
%				%\cmidrule(lr){2-2}
%				%\cmidrule(lr){3-6}		
%				\midrule
%				&& 3D CTA & age & gender & CHD & PAD \\
%				\cmidrule(lr){3-3}
%				\cmidrule(lr){4-4}
%				\cmidrule(lr){5-5}
%				\cmidrule(lr){6-6}
%				\cmidrule(lr){7-7}
%				ResNetMLP &from scratch&84.65\%&1.29\%&5.96\%&0.00\%&8.10\%\\
%				& pretrained temp. frozen & 82.40\% & 1.42\% &8.50\% &0.00\% & 7.68\%\\
%				\addlinespace[2mm]
%				ViTMLP &from scratch&0.14\%&4.66\%&73.72\%&0.00\% &21.48\%\\
%				& pretrained frozen &54.28\% &1.87\% &35.94\% &0.00\% &7.91\% \\
%				\bottomrule
%			\end{tabular}
%		\end{table}
		
		%\begin{figure}[h!]
		%	\centering
		%	\includegraphics[width=0.6\textwidth]{loss_vs_epoch.pdf}
		%	\caption{Training loss over epochs for all models.}
		%	\label{fig:loss_plot}
		%\end{figure}
		
		%\let\clearpage\relax
		
	\end{document}